\documentclass{SIP}
\usepackage{epstopdf, epsfig} 
\usepackage{multirow}
\usepackage{pifont}
\newcommand{\cmark}{\ding{51}}%
\newcommand{\xmark}{\ding{55}}%

\usepackage{xspace}
\usepackage{times}
\usepackage{epsfig}
\usepackage{graphicx}
\usepackage{amsmath}
\usepackage{amssymb}
\usepackage{mathrsfs}
\usepackage{enumitem}
\usepackage{subfigure}
\usepackage{array}
\usepackage{tabularx}
\usepackage{multirow}
\usepackage{multicol}

\usepackage[pagebackref=true,breaklinks=true,letterpaper=true,colorlinks,bookmarks=false]{hyperref}

\begin{document}
\title[Resolving Semantic Gap]{Bridging Gap between Image Pixels and 
Semantics via Supervision: A Survey}
\author[Jiali Duan, \textit{et al}.]{Jiali Duan$^{1}$, C.-C. Jay Kuo$^{1}$}
\address{\add{1}{University of Southern California, Los Angeles, California, United States}}
\corres{\name{Jiali Duan}
\email{jialidua@usc.edu}}

\begin{abstract}

The fact that there exists a gap between low-level features and semantic
meanings of images, called the semantic gap, is known for decades.
Resolution of the semantic gap is a long standing problem.  The semantic
gap problem is reviewed and a survey on recent efforts in bridging the
gap is made in this work. Most importantly, we claim that the semantic
gap is primarily bridged through supervised learning today.  Experiences
are drawn from two application domains to illustrate this point: 1)
object detection and 2) metric learning for content-based image
retrieval (CBIR). To begin with, this paper offers a historical
retrospective on supervision, makes a gradual transition to the modern
data-driven methodology and introduces commonly used datasets.  Then, it
summarizes various supervision methods to bridge the semantic gap in the
context of object detection and metric learning. 

\end{abstract}

\keywords{Semantic Gap, Semantic Understanding, Content-based Image
Retrieval, Supervision, Object Detection, Metric Learning.}

\maketitle

\section{Introduction}\label{sec:introduction}

Computer vision deals with how computers can gain high-level
understanding of visual contents, which are represented by pixels.
High-level understanding of visual inputs demands the capability to
learn the semantics conveyed through raw pixels.  The fact that there
exists a gap between low-level features and semantic meanings of images,
is known for decades.  It is the consequence of ``lack of coincidence
between the information that one can extract from the visual data and
the interpretation that the same data have for a user in a given
situation'' \cite{smeulders2000content}.  It is well known as the
semantic gap. 

There is a growing awareness in the computer vision community that the
key to today's vision problems lies in resolving the gap between image
pixels and semantics. There has been a substantial amount of progress in
bridging the gap in recent years.  In our opinion, this advancement is
primarily attributed to supervised learning and our survey paper is
written around this central idea.  Supervision manifests itself through
two aspects: 1) large-scale, high-quality annotated data, and 2)
well-designed optimization objectives. The two aspects often come into
play synergistically. For example, the design of optimization objectives
highly depend on annotations.  The optimization procedure often entails
a minimum amount of labeled data and it is expected to scale well with
more data. 

Fig.~\ref{fig:intro_cover} gives an example of how supervision in the
form of ``annotated data" advances the object classification field.  The
figure shows the progress of top-1 classification accuracy as a function
of time with respect to the ImageNet dataset. Undoubtedly, the
large-scale annotated ImageNet dataset contributes significantly to
semantic image understanding. Thus, by associating the progress in
bridging the semantic gap with the construction of large-scale annotated
datasets, we will introduce several commonly used datasets that help
provide supervision to capture the semantic information. 

The second aspect, optimization design under supervision, is the main
focus of this paper. Although there is a vast amount of references on
this topic, our goal is to shed light on the role of supervision in
bridging the semantic gap. To illustrate this point, we choose two
representative domains for elaboration: 1) object detection and 2)
metric learning in the context of content-based image retrieval (CBIR).
Both are fundamental computer vision problems. We summarize various
forms of supervision used to bridge the semantic gap in the two fields,
including fully-supervised learning, semi-supervised learning,
weakly-supervised learning, self-supervised learning, etc. 

\begin{figure*}[h]
\centering
\includegraphics[width=0.8\linewidth]{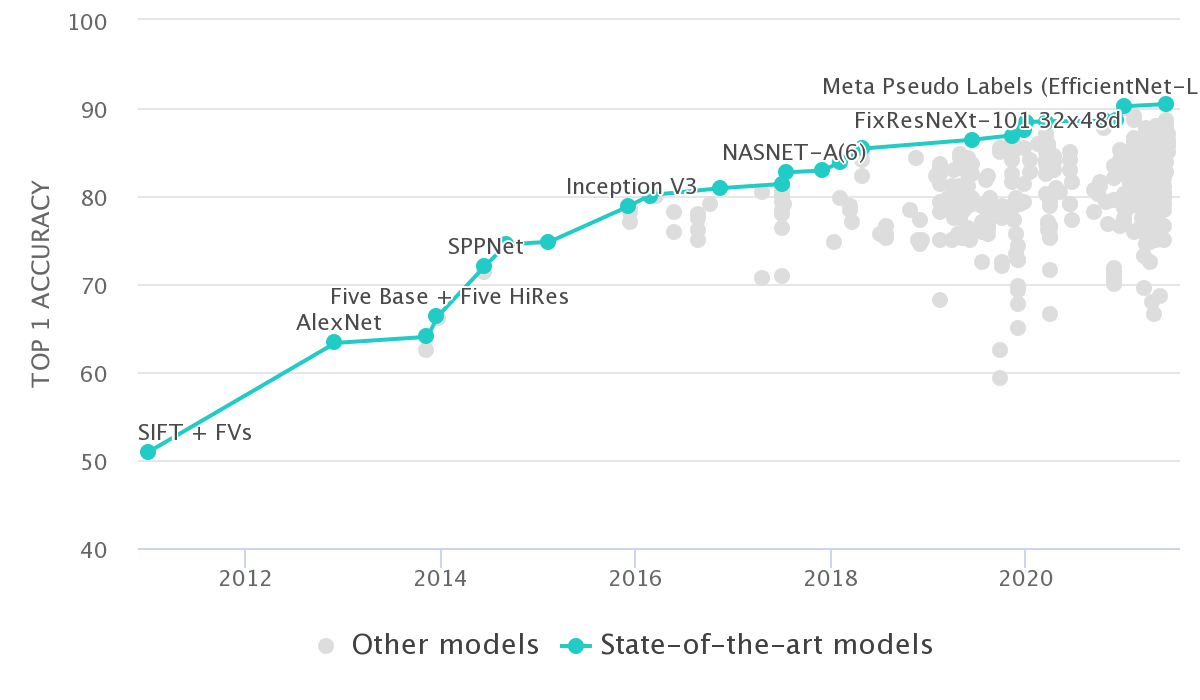}
\end{figure*}

\begin{table*}[h]
\centering
\begin{tabular}{|l|c|c|c|c|c|}
\hline
 Method & Top 1 ACC (\%)& Top 5 ACC (\%) & Params (M) & Extra Data & Year  \\
 Meta Pseudo labels~\cite{pham2021meta} & 90.35  & 98.8 & 480  & \cmark &  2020\\
 FixResNeXt-101 32x48d~\cite{touvron2019fixing} & 86.4  & 98.0 & 829 & \cmark & 2019  \\
 NASNET-A(6)~\cite{zoph2018learning} & 82.7 & 96.2 & 88.9 & \xmark & 2017 \\
 Inception-V3~\cite{szegedy2016rethinking} & 78.8 & - & - & \xmark & 2015 \\
 SPPNet~\cite{he2015spatial} & 72.14 & 91.86 & - & \cmark & 2014 \\
 Five Base + Five HiRes~\cite{howard2013some} & 66.3 & 86.3 & - & \xmark & 2013 \\
 AlexNet~\cite{krizhevsky2012imagenet} & 63.3 & 84.6 & 60 & \xmark & 2012 \\
 SIFT + FVs~\cite{ke2004pca} & 50.9 & 73.8 & - & \xmark & 2010 \\
 \hline
\end{tabular}

\vspace{2mm}
\caption{
The top 1 classification accuracy for ImageNet as a function
of years \cite{pwc2021imagenet}. The ImageNet is an important dataset that drives research on
object classification/recognition, and the associated image labels offer
supervision to address the semantic gap problem.}\label{fig:intro_cover}
\end{table*}

{\bf Semantic Gap.} Understanding semantics is the most fundamental step
in all kinds of computer vision problems as it paves the way for general
artificial intelligence. The semantic gap is
a general term widely used in content-based image retrieval (CBIR).  It
is defined in~\cite{rui1999image} as follows. ``Humans tend to use high-level concepts
in everyday life. However, what current computer vision techniques can
automatically extract from image are mostly low-level features. In
constrained applications, such as the human face and finger print, it is
possible to link the low-level features to high-level concepts (faces or
finger prints). In a general setting, however, the low-level features do
not have a direct link to the high-level concepts."

\begin{figure*}[thb]
\centering
\includegraphics[width=1.0\linewidth]{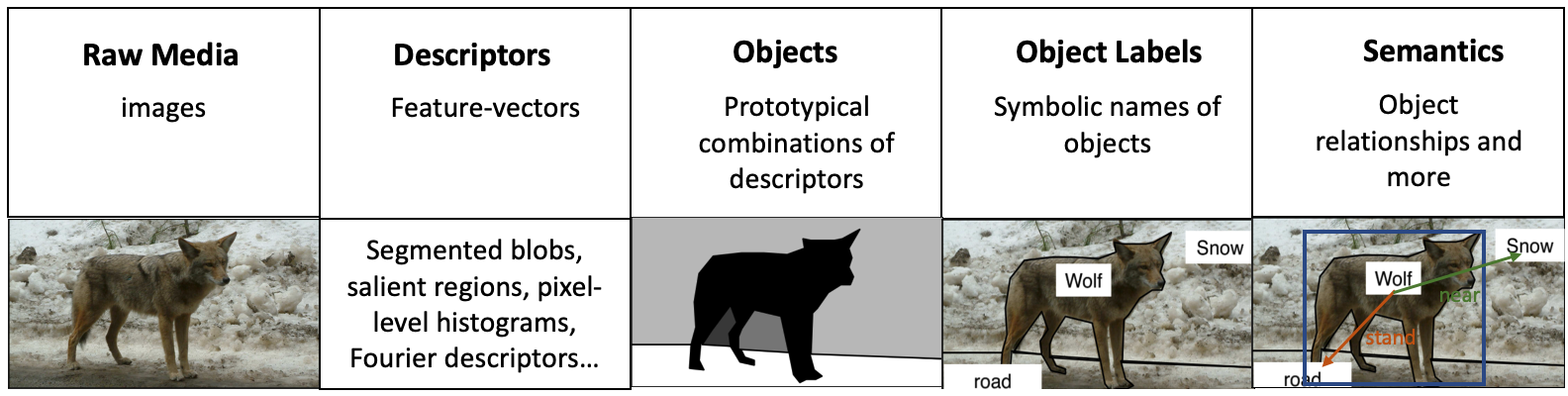}
\caption{Illustration of the semantic gap in multiple representation levels between raw pixels and full semantics~\cite{hare2006mind}. Left to right indicates increasing levels of semantic understanding, based on information of the previous step. For example, objects are agglomeration of feature descriptors and object labels are derived based on features for the objects. The rightmost is closest to human-level understanding of the input such as object relationships and more. }\label{fig:semantic-gap}
\end{figure*}

The semantic gap manifests itself through
different semantic understanding levels as shown in
Fig.~\ref{fig:semantic-gap}.  The raw media representation lies at the
lowest level. In the context of object detection and image retrieval,
the basic representation unit is the RGB pixel. At a higher level,
low-level feature vectors are extracted by image analysis tools. This
process is sometimes called low-level computer vision. The extracted
features can be in form of segmented blobs, texture statistics, simple
colour histograms, and other hand-crafted feature vectors used to
represent parts or full images. As these feature descriptors are often
human-engineered, they may require the domain knowledge from experts.
At the next higher level, there are object representations which may be
prototypical combinations of feature vectors or other more explicit
representations. Once identified, objects are given symbolic labels such
as object names.  Labels may be general or specific, e.g., an animal or
a wolf. Labelling all objects does not necessarily capture the full
semantics of an image since there may exist relationship between
objects. Furthermore, the amount of labor required by labeling is
tremendous. At the highest level as shown in
Fig.~\ref{fig:semantic-gap}, we target at understanding the relations
between objects and the holistic meaning of an image. 

Semantics is a broad topic and in this survey, we choose object detection and metric learning as two examples for the following reasons. 
\begin{enumerate}
    \item We focus on resolving the gap
between image pixels and semantic meanings of images, also known as the
semantic gap. Although there are other vision applications such as scene graph generation~\cite{xu2017scene}, visual question answering~\cite{anderson2018bottom}, human-object interaction~\cite{yao2010modeling} that require richer semantic understanding, they are higher-level computer vision tasks that build upon object detection and metric learning. The two topics are fundamental and closer to the pixel level, where semantics is hard to capture. 
    \item  Another reason is that there is a decent accumulation of various
supervisions for the two applications, making them good choices to study
the role that supervision plays in bridging the semantic gap. 
\end{enumerate}
That being said, there are many other useful tools such as semantic parsing~\cite{berant2013semantic} in the NLP domain that help understand the semantics.

{\bf Comparison with Previous Reviews.} Many papers that aimed to study
the semantic gap problem have been published.  They are summarized in
Table~\ref{tab:semantic_surveys}.  The list includes many excellent
surveys on the specific problem of image
retrieval~\cite{alzubaidi2017new, chen2012ilike, ma2010bridging,
hare2006mind}, video retrieval~\cite{li2004bridging}, semantic
segmentation~\cite{pang2019towards}, etc. 

Recent success and dominance of deep learning based
methods uphold the promise to achieve this goal. To this end, there have
been many published surveys on deep learning such as the work of Bengio
{\em et al.}~\cite{bengio2013representation}, LeCun {\em et al.}
\cite{lecun2015deep}, Gu {\em et al.}~\cite{gu2018recent}, and recent
tutorials given at CVPR and ICCV. Although deep learning based methods
have been proposed to bridge the semantic gap, we are unaware of any
comprehensive survey that attempts to unify them at a higher level. In
this survey, we organize papers and summarize their ideas by grouping
them into different supervision forms such as fully-supervised,
unsupervised, semi-supervised, self-supervised and weak-supervised etc. 
Another important distinction between our paper and previous ones is
that we do not restrict ourselves to a specific problem but focus on
resolving the semantic gap at a broader context.

\begin{table*}[!t]
\caption {Summary of surveys on the semantic gap study}\label{tab:semantic_surveys}
\centering
\renewcommand{\arraystretch}{1.5}
\setlength\arrayrulewidth{0.2mm}
\setlength\tabcolsep{2pt}
\resizebox*{16cm}{!}{
\begin{tabular}{!{\vrule width1.2bp}c|p{6cm}<{\centering}|c|c|c|p{9cm}<
{\centering}!{\vrule width1.2bp}} \hline
\footnotesize No.   & \footnotesize Survey Title  & \footnotesize Ref.	 & 
\footnotesize Year & \footnotesize Venue	 & \footnotesize Content  \\ \hline
\raisebox{-1.5ex}[0pt]{\footnotesize 1 }& \footnotesize Bridging the semantic gap 
in image retrieval  &\raisebox{-1.5ex}[0pt]{ \footnotesize \cite{zhao2002bridging} 	}
&\raisebox{-1.5ex}[0pt]{ \footnotesize 2002}
& \raisebox{-1.5ex}[0pt]{\footnotesize IGI}	& \footnotesize image retrieval \\ \hline
\raisebox{-1.5ex}[0pt]{\footnotesize 2 }& \footnotesize Bridging the semantic 
gap in sports video retrieval and summarization & \raisebox{-1.5ex}[0pt]{\footnotesize 
\cite{li2004bridging}}	 &\raisebox{-1.5ex}[0pt]{ \footnotesize  2004 }
& \raisebox{-1.5ex}[0pt]{\footnotesize JVCI }& \footnotesize \raisebox{-1.5ex}
[0pt]{	sports video retrieval} \\ \hline
\raisebox{-1.5ex}[0pt]{\footnotesize 3} & \footnotesize	  Towards bridging the semantic gap 
in multimedia annotation and retrieval	 & \raisebox{-1.5ex}[0pt]{\footnotesize \cite{vembu2006towards} }
& \raisebox{-1.5ex}[0pt]{\footnotesize	2006	}& \raisebox{-1.5ex}[0pt]{\footnotesize SWAMM}
& \footnotesize	multimedia retrieval \\ \hline
\footnotesize 4 & \footnotesize Foafing the music: Bridging the semantic gap in music recommendation	 
& \footnotesize \cite{celma2008foafing} & \footnotesize 2008 & \footnotesize ISWC & \footnotesize	 
music recommendation \\ \hline
\footnotesize 5 & \footnotesize Mind the Gap: Another look at the problem of the semantic gap in 
image retrieval & \footnotesize \cite{hare2006mind} & \footnotesize 2006 & \footnotesize ISOP & 
\footnotesize image-retrieval \\ \hline
\raisebox{-1.5ex}[0pt]{\footnotesize 6	} & \footnotesize  Bridging the semantic gap in 
multimedia information retrieval: Top-down and bottom-up approaches	& 
\raisebox{-1.5ex}[0pt]{\footnotesize \cite{hare2006bridging}}
& \raisebox{-1.5ex}[0pt]{\footnotesize 2015 }& \raisebox{-1.5ex}[0pt]{ \footnotesize -}& 
\raisebox{-1.5ex}[0pt]{ \footnotesize multimedia information retrieval} \\ \hline
\footnotesize 7 & \footnotesize Bridging the semantic gap for texture-based image retrieval 
and navigation & \footnotesize \cite{idrissi2009bridging} & \footnotesize 2009	& \footnotesize JOM	
& \footnotesize image-retrieval \\ \hline
\footnotesize 8	 & \footnotesize Bridging the semantic gap between image contents and 
tags & \footnotesize \cite{ma2010bridging} & \footnotesize 2010	& \footnotesize 
IEEE MultiMedia	& \footnotesize image retrieval \\ \hline
\raisebox{-1.5ex}[0pt]{\footnotesize 9} &
\footnotesize ilike: Bridging the semantic gap in vertical image search by integrating text 
and visual features &\raisebox{-1.5ex}[0pt]{ \footnotesize \cite{chen2012ilike}}	 
&\raisebox{-1.5ex}[0pt]{ \footnotesize	2012}
& \raisebox{-1.5ex}[0pt]{\footnotesize 	KDE}	& \footnotesize image-text retrieval \\ \hline
\raisebox{-1.5ex}[0pt]{ \footnotesize	10 }& \footnotesize A new strategy for bridging 
the semantic gap in image retrieval	& \raisebox{-1.5ex}[0pt]{\footnotesize \cite{alzubaidi2017new}}
&\raisebox{-1.5ex}[0pt]{ \footnotesize 2017}&\raisebox{-1.5ex}[0pt]{ \footnotesize	JCSE }  
& \raisebox{-1.5ex}[0pt]{ \footnotesize	image-retrieval}  \\ \hline
\raisebox{-1.5ex}[0pt]{ \footnotesize	11 }& \footnotesize Towards bridging semantic gap to 
improve semantic segmentation	& \raisebox{-1.5ex}[0pt]{\footnotesize \cite{pang2019towards}}
&\raisebox{-1.5ex}[0pt]{ \footnotesize 2019}&\raisebox{-1.5ex}[0pt]{ \footnotesize	ICCV}  
& \raisebox{-1.5ex}[0pt]{ \footnotesize	semantic segmentation}  \\ \hline
\end{tabular}
}
\end{table*}

\begin{table*}[!t]
\caption {Summary of surveys on object detection.}\label{tab:object_detection}
\centering
\renewcommand{\arraystretch}{1.5}
\setlength\arrayrulewidth{0.2mm}
\setlength\tabcolsep{2pt}
\resizebox*{16cm}{!}{
\begin{tabular}{!{\vrule width1.2bp}c|p{6cm}<{\centering}|c|c|c|p{9cm}<{\centering}!
{\vrule width1.2bp}} \hline
\footnotesize No.   & \footnotesize Survey Title  & \footnotesize Ref.	 
& \footnotesize Year & \footnotesize Venue	 & \footnotesize Content  \\ \hline
\raisebox{-1.5ex}[0pt]{\footnotesize 1 }& \footnotesize 	 Survey of Pedestrian 
Detection for Advanced Driver Assistance Systems & \raisebox{-1.5ex}[0pt]{\footnotesize 
\cite{geronimo2009survey}}	 &\raisebox{-1.5ex}[0pt]{ \footnotesize  2010 }
& \raisebox{-1.5ex}[0pt]{\footnotesize PAMI }& \footnotesize \raisebox{-1.5ex}[0pt]{	 
pedestrian detection} \\  \hline
\footnotesize 2 & \footnotesize 	Detecting Faces in Images: A Survey	 & 
\footnotesize \cite{yang2002detecting} & \footnotesize 2002 & \footnotesize PAMI & \footnotesize	 
First survey of face detection from a single image   \\ \hline
\raisebox{-1.5ex}[0pt]{\footnotesize 3	} & \footnotesize A Survey on Face Detection in the
Wild: Past, Present and Future	& \raisebox{-1.5ex}[0pt]{\footnotesize \cite{zafeiriou2015survey}}
& \raisebox{-1.5ex}[0pt]{\footnotesize 2015 }& \raisebox{-1.5ex}[0pt]{ \footnotesize CVIU}& 
\raisebox{-1.5ex}[0pt]{ \footnotesize face dtection} \\ \hline
\raisebox{-1.5ex}[0pt]{\footnotesize 4	}& \raisebox{-1.5ex}[0pt]{
\footnotesize On Road Vehicle Detection: A Review}	 &\raisebox{-1.5ex}[0pt]{ 
\footnotesize \cite{sun2006road} }&\raisebox{-1.5ex}[0pt]{ \footnotesize	2006	}&
\raisebox{-1.5ex}[0pt]{ \footnotesize PAMI}	& \footnotesize vehicle detection  \\ \hline
\footnotesize 5	 & \footnotesize Text Detection and Recognition in Imagery: 
A Survey & \footnotesize \cite{ye2014text} & \footnotesize 2015	& \footnotesize 
PAMI	& \footnotesize text detection \\ \hline
\footnotesize 6	& \footnotesize Object Class Detection: A Survey	& \footnotesize 
\cite{zhang2013object} & \footnotesize	 2013	& \footnotesize ACM CS & \footnotesize	 
object detection before 2011  \\ \hline
\footnotesize 7	& \footnotesize Salient Object Detection: A Survey		 
& \footnotesize \cite{borji2019salient} & \footnotesize2014	& \footnotesize arXiv	
& \footnotesize A survey for salient object detection \\ \hline
\raisebox{-1.5ex}[0pt]{ \footnotesize 8 }&
\footnotesize	A Survey on Deep Learning in Medical Image Analysis& \raisebox{-1.5ex}[0pt]{
\footnotesize \cite{litjens2017survey} }&\raisebox{-1.5ex}[0pt]{ \footnotesize	 2017}
&\raisebox{-1.5ex}[0pt]{ \footnotesize 	MIA}	& \footnotesize object detection for medical 
images  \\ \hline
\footnotesize {9}& \footnotesize	 Deep Learning for Generic Object Detection: A Survey 
& \footnotesize \cite{liu2020deep} & \footnotesize 2020	& \footnotesize JVCI	 & \footnotesize 
A comprehensive survey of deep learning for generic object detection \\ \hline
\end{tabular}}
\end{table*}

\begin{table*}[!t]
\caption {Summary of surveys on metric learning.}\label{tab:metric_learning}
\centering
\renewcommand{\arraystretch}{1.5}
\setlength\arrayrulewidth{0.2mm}
\setlength\tabcolsep{2pt}
\resizebox*{16cm}{!}{
\begin{tabular}{!{\vrule width1.2bp}c|p{6cm}<{\centering}|c|c|c|p{9cm}<{\centering}!{\vrule width1.2bp}}
\hline
\footnotesize No.   & \footnotesize Survey Title  & \footnotesize Ref.	 & \footnotesize Year & 
\footnotesize Venue	 & \footnotesize Content  \\ \hline
\raisebox{-1.5ex}[0pt]{\footnotesize 1 }& \footnotesize 	 Metric learning: A survey. & 
\raisebox{-1.5ex}[0pt]{\footnotesize \cite{kulis2012metric}}	 &\raisebox{-1.5ex}[0pt]{ 
\footnotesize  2012 } & \raisebox{-1.5ex}[0pt]{\footnotesize Journal }& \footnotesize 
\raisebox{-1.5ex}[0pt]{metric learning based on hand-crafted features} \\ \hline
\footnotesize 2 & \footnotesize Distance metric learning: A comprehensive survey & \footnotesize 
\cite{yang2006distance} & \footnotesize 2006 & \footnotesize MSU & \footnotesize	 
traditional metric learning methods before 2006   \\ \hline
\raisebox{-1.5ex}[0pt]{\footnotesize 3	} & \footnotesize Deep metric learning: A survey	
& \raisebox{-1.5ex}[0pt]{\footnotesize \cite{kaya2019deep}}
& \raisebox{-1.5ex}[0pt]{\footnotesize 2019 }& \raisebox{-1.5ex}[0pt]{ \footnotesize Symmetry}
& \raisebox{-1.5ex}[0pt]{ \footnotesize deep learning based metric learning before 2019} \\ \hline
\raisebox{-1.5ex}[0pt]{\footnotesize 4	}& 
\footnotesize A survey on metric learning for feature vectors and structured data	 
&\raisebox{-1.5ex}[0pt]{ \footnotesize \cite{bellet2013survey} }&\raisebox{-1.5ex}[0pt]{ 
\footnotesize	2013	}& \raisebox{-1.5ex}[0pt]{ \footnotesize arXiv}	& \footnotesize 
traditional metric learning methods  \\ \hline
\footnotesize 5	 & \footnotesize Survey on distance metric learning and dimensionality 
reduction in data mining & \footnotesize \cite{wang2015survey} & \footnotesize 2015	& 
\footnotesize DMKD	& \footnotesize metric learning in data mining \\ \hline
\footnotesize 6	& \footnotesize Survey and experimental study on metric learning methods & 
\footnotesize \cite{li2018survey} & \footnotesize	 2018	& \footnotesize Neural networks & 
\footnotesize	 experimental benchmark for metric learning  \\ \hline
\footnotesize 7	& \footnotesize An overview of distance metric learning		 & 
\footnotesize \cite{yang2007overview} & \footnotesize2017	& \footnotesize CVPR	& 
\footnotesize overview of traditional metric learning methods \\ \hline
\raisebox{-1.5ex}[0pt]{ \footnotesize 8 }&
\footnotesize	A decade survey of content based image retrieval using deep learning  &
\raisebox{-1.5ex}[0pt]{\footnotesize \cite{dubey2021decade} } &\raisebox{-1.5ex}[0pt]{ 
\footnotesize	 2021}  &\raisebox{-1.5ex}[0pt]{ \footnotesize CSVT}	& \footnotesize 
deep learning based methods for metric learning in recent 10 years  \\ \hline
\footnotesize {9}& \footnotesize A metric learning reality check & \footnotesize 
\cite{musgrave2020metric} & \footnotesize 2020	& \footnotesize ECCV	 & 
\footnotesize A benchmark for DML \\ \hline
\end{tabular}
}
\end{table*}

{\bf Scope of our work.} The central theme of our paper is supervision,
which we believe is the key to semantic gap resolution.  However,
supervision is a broad topic and we need to limit our scope to two
important problems (i.e., object detection and metric learning) as they
help reveal our insights. There are still too many papers on these two
topics, and compiling an extensive list of the state-of-the-art methods
of both is beyond the scope of a paper of reasonable length.  Instead,
for domain-specific surveys, readers are referred to
Table~\ref{tab:object_detection} and Table~\ref{tab:metric_learning},
respectively. 

\begin{figure}[h]
\centering
\includegraphics[width=1.0\linewidth]{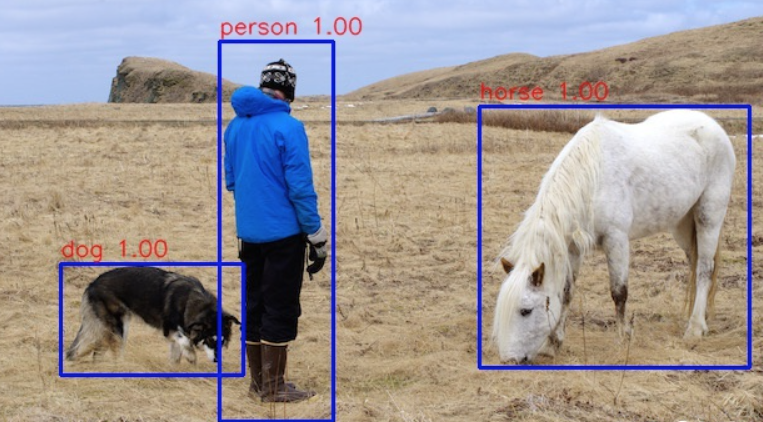}
\caption{An object detection example.}\label{fig:detection_cover}
\vspace{-2mm}
\end{figure}

The first selected illustrative topic is object detection. It is a
fundamental computer vision problem and serves as a building block for
image segmentation~\cite{haralick1985image}, object
tracking~\cite{chen2019fast, yan2021learning,li2019siamvgg,
zhang2020ocean}, landmark detection~\cite{zadeh2017convolutional,
xu2020anchorface}, etc.  Its goal is to identify the location (i.e., the
coordinates of a bounding box) of an object instance and its
corresponding category (e.g., persons, pedestrians, cars, animals).
Previous survey papers have covered different aspects of object
detection such as pedestrian detection~\cite{enzweiler2008monocular,
geronimo2009survey,dollar2011pedestrian}, face
detection~\cite{yang2002detecting, zafeiriou2015survey,duan2016face}, vehicle
detection~\cite{sun2006road}, gesture recognition~\cite{duan2018unified,duan2016multi}, text detection~\cite{ye2014text}, etc.
There are also a number of review papers on generic object detection
methods~\cite{grauman2011visual, andreopoulos201350, zhang2013object,
liu2020deep}.  Here, our goal is to provide a connection between object
detection and different supervisions.  An example of object detection is
given in Fig.~\ref{fig:detection_cover}. In a fully-supervised setting,
object classes and their bounding boxes are annotated in each image.
However, it is expensive and often impossible to manually labor all
possible objects in the real world. That is the reason other forms of
supervision have to be developed. We will review various supervisions
developed for object detection in Sec. \ref{sec:obj}.

The second exemplary topic is metric learning for image retrieval. It
learns a distance metric so as to establish similarity or dissimilarity
between objects and find applications in image, video and multimedia
retrieval and music recommendation. While metric learning aims to
reduce the distance between similar objects, it also intends to increase
the distance between dissimilar objects. Typically, deep metric learning requires
the class label for each individual sample. This demanding requirement
prohibits its applicability in wild scenarios. Efforts have been made to
relax the stringent requirement so as to accommodate other learning
environments such as semi-supervised, weakly-supervised,
pseudo-supervised, self-supervised and even unsupervised approaches.  We
will review deep metric learning methods with different supervision
types in Sec. \ref{sec:cbir}. 

\section{Background Review}\label{sec:perspective}

{\bf Object Detection.} Different from current deep learning based
methods which extract the feature representation from images implicitly
and automatically, traditional object detection methods rely heavily on
hand-crafted features. A traditional object detection pipeline consists
of the following three steps. 
\begin{enumerate}
\item Extract a region of interest (say, a region that enclose objects).
\item Obtain features from the region of interest. They are often handcrafted 
based on the domain knowledge. 
\item Classify the region of interest into a certain object class based on
extracted features. 
\end{enumerate}

In the first step, regions of interest (ROIs) are often extracted with a
sliding window approach. It requires the choice of certain
hyper-parameters such as window's width, height, stride and aspect
ratio. As the number of objects in the scene increases, this brute force
enumeration approach can lead to a very high computational cost. Later,
researchers came up with efficient yet heuristic approaches such as
selective search~\cite{uijlings2013selective}, edge boxes~
\cite{zitnick2014edge}, box refinement~\cite{li2017box} to generate
region proposals.  The second step involves feature engineering that
plays a crucial role in the performance. One seminal work is the Scale
Invariant Feature Transform (SIFT)~\cite{lowe1999object}. It was
designed to be robust against changes in translation, scale, rotation,
illumination, viewpoint and occlusion. Other local representative
descriptors~\cite{mikolajczyk2005performance} include Haar-like
features~\cite{viola2001rapid}, local binary patterns
(LBP)~\cite{ojala2002multiresolution} and region
covariances~\cite{tuzel2006region}. The histogram of oriented gradients
(HOGs)~\cite{dalal2005histograms} is an important improvement over SIFT
and offers a better object descriptor. The HOG feature is robust against
local deformation and illumination, and it has been widely used in
classical object detectors.  The last step is classification based on
features of each ROI. Most commonly used classifiers include the support
vector machine (SVM)~ \cite{cortes1995support},
AdaBoost~\cite{freund1997decision} and random forest
(RF)~\cite{svetnik2003random}. 

One famous example of the three-step pipeline is the Viola-Jones face
detector~\cite{viola2001rapid}. It adopts a sliding window approach to
check if a face object is included in the window. To improve the
detection speed, it uses an Adaboost training approach and cascades
classifiers to improve the detection performance. The deformable part
model (DPM)~ \cite{felzenszwalb2008discriminatively} offers another
milestone in the traditional object detection framework.  It consists of
a root-filter and multiple part-filters. DPM improves HOG using hard
negative mining, bounding box regression and context priming.  It was
the champion solution in the Pascal-VOC Challenge from
2007-2009~\cite{everingham2010pascal}. The cascaded pipelines of
``hand-crafted feature description'' followed by ``discriminative
classification'' dominated many computer vision tasks, including object
detection, for years. Even with significant advancement, there is a
substantial gap between the classical object detector and human
recognition capability.  This gap is attributed to two main barriers:
the limited representation capability of hand-crafted features and lack
of sufficient supervision.  Deep learning can learn powerful features
automatically to overcome the first barrier. The construction of larger
and larger labeled datasets addresses the second barrier. 

{\bf Metric Learning.} Metric learning is a branch of machine learning.
It learns a distance metric that establishes similarity and
dissimilarity between objects from training images. The objective is to
reduce the distance between similar objects while increasing that
between dissimilar ones. The task is also known as similarity learning
and it is most commonly used in image retrieval~\cite{hoi2010semi,
hsieh2017collaborative, lee2008rank, mcfee2010metric, yang2008boosting},
person re-identification~\cite{yi2014deep, liao2015person,
xiong2014person} and face verification~\cite{nguyen2010cosine}, etc. 

For a given query image, a content-based image retrieval (CBIR) system
\cite{rui1999image} return a ranked list of images from the database
based on a similarity measure between the query and retrieved images.
CBIR is a challenging problem since it is often that many (or even all)
of those returned images do not look similar to the query one from a
human perspective. This is because that most similarity metrics refer to
distances of low-level features. They do not correlate well with
semantic similarity perceived by humans.

Traditional CBIR research focused on two areas: feature design and
distance metric selection. Research on the application of hand-crafted
features to CBIR was rich. SIFT~\cite{lowe1999object} and
LBP~\cite{ojala2002multiresolution} were widely used features. A
histogram-based similarity measure was proposed in \cite{swain1991color}
for image retrieval. The K-means clustering approach was used in
\cite{sethi2001mining} to discover the patterns of data in low-level
feature space using the color information.  A nonlinear mapping approach
based on sparse kernel learning was studied in \cite{moran2014sparse}.
Other features were designed based on the prior knowledge and domain
expertise for specific application. For example,
LOMO~\cite{liao2015person,duan2016benchmark} was developed to deal with illumination and
viewpoint changes for the matching of person images.  As to distance
metrics, common choices include Euclidean, Mahalanobis
\cite{chopra2005learning,davis2007information,weinberger2009distance},
and Kullback-Leibler distances. Higher performance may be achieved by
mapping the problem to a non-linear space through kernel methods.  These
nonlinear methods are often used in combinationn of regularization
techniques~\cite{jin2009regularized} to avoid overfitting. 

{\bf Large-Scale Labeled Datasets.} Resolution of the semantic gap has
been a long standing problem.  There is a substantial progress on this
topic in the last decade. This progress is attributed to the realization
of the importance of supervised learning. 

While the semantic gap is bridged by supervision, the key to supervised
learning is the availability of large-scale human annotated datasets.
Judged by today's standard, the sizes of labeled datasets were quite
small before 2010. This practice can be attributed to several reasons.
First, it is about the labeling cost. The labor required to annotate
collected data is substantial. Second, most traditional methods do not
scale well with the data size. When most solutions do not work well for
small datasets, the motivation in building larger datasets would not be
strong. Third, since people can understand semantic meanings from a
small set of examples (i.e. weakly-supervised learning), it is natural
to expect powerful vision algorithms to do the same. For all these
reasons, the importance of ``large-scale supersion'' was not appreciated
and practiced until the last decade. 

The situation began to change with the introduction of the ImageNet
dataset~\cite{deng2009imagenet}, which was viewed as the engine to drive
deep learning in early 10s. That is, deep learning has gained widespread
attention and popularity after Krizhevsky et
al.~\cite{krizhevsky2012imagenet} achieved record-breaking image
classification accuracy in the Large Scale Visual Recognition Challenge
(ILSVRC)~\cite{russakovsky2015imagenet} in 2012. Although deep learning
provides a mechanism in extracting powerful representative features,
feature extraction is not the main objective of deep learning but a
byproduct. Deep learning relies heavily on supervision. It attempts to
build a mapping from images to labels by certain neural networks. In
other words, it uses human labels as the ground truth and provides a
nonlinear mechanism that minimizes the error between the predicted and
true labels. 

\begin{table*}[!t]
\caption {Commonly used databases for object detection.} \label{tab:obj_data}
\centering
\renewcommand{\arraystretch}{1.2}
\setlength\arrayrulewidth{0.2mm}
\setlength\tabcolsep{1pt}
\resizebox*{18cm}{!}{
\begin{tabular}{!{\vrule width1.2bp}c|c|c|c|c|c|c|p{8cm}!{\vrule width1.2bp}}
\hline
 \scriptsize   \shortstack [c] {\textbf{Dataset}\\ \textbf{Name}} & \scriptsize   \shortstack [c]
{\textbf{Total} \\ \textbf{Images}} & \scriptsize   \shortstack [c]
{\textbf{Categories}} & \scriptsize   \shortstack [c]
{\textbf{Images Per} \\ \textbf{Category}} & \scriptsize   \shortstack [c]
{\textbf{Objects Per }\\ \textbf{Image}} & \scriptsize   \shortstack [c] {\textbf{Image} \\ \textbf{Size} }
& \scriptsize   \shortstack [c] {\textbf{Started} \\  \textbf{Year}}
&\raisebox{1.3ex}[0pt]{ \scriptsize   \shortstack [c] {$\quad\quad\quad\quad\quad\quad\quad\quad\quad\quad\quad\quad$\textbf{Highlights}}} \\
\hline
\raisebox{-6.3ex}[0pt]{\scriptsize \shortstack [c] {PASCAL \\VOC \\ (2012)} } &\raisebox{-5ex}[0pt]{\scriptsize   $11,540$ }& \raisebox{-5ex}[0pt]{\scriptsize $20$} & \raisebox{-5ex}[0pt]{ \scriptsize  $303\sim4087$}& \raisebox{-5ex}[0pt]{\scriptsize $2.4$}
& \raisebox{-5ex}[0pt]{\scriptsize $470\times380$}& \raisebox{-5ex}[0pt]{ \scriptsize  $2005$} & \scriptsize  Covers only 20 categories that are common in everyday life; Large number of training images; Close to real-world applications; Significantly larger intraclass variations; Objects in scene context; Multiple objects in one image; Contains many difficult samples.  \\
\hline
\raisebox{-3.3ex}[0pt]{\scriptsize ImageNet} & \raisebox{-3.3ex}[0pt]{\scriptsize   $14M$ } &
\raisebox{-3.3ex}[0pt]{ \scriptsize $21,841$} &\raisebox{-3.3ex}[0pt]{ \scriptsize  $-$}
&\raisebox{-3.3ex}[0pt]{ \scriptsize  $1.5$}
& \raisebox{-3.3ex}[0pt]{\scriptsize $500\times400$}&\raisebox{-3.3ex}[0pt]{ \scriptsize  $2009$} & \scriptsize  Large number of object categories; More instances and more categories of objects per image; More challenging than PASCAL VOC; Backbone of the ILSVRC challenge; Images are object-centric. \\
\hline
\raisebox{-3.3ex}[0pt]{\scriptsize MS COCO} & \raisebox{-3.3ex}[0pt]{\scriptsize   $328,000+$ } & \raisebox{-3.3ex}[0pt]{\scriptsize $91$} & \raisebox{-3.3ex}[0pt]{ \scriptsize  $-$}
 & \raisebox{-3.3ex}[0pt]{ \scriptsize  $7.3$}
& \raisebox{-3.3ex}[0pt]{ \scriptsize $640\times480$}& \raisebox{-3.3ex}[0pt]{\scriptsize  $2014$} & \scriptsize  Even closer to real world scenarios; Each image contains more instances of objects and richer object annotation information; Contains object segmentation notation data that is not available in the ImageNet dataset. \\
\hline
\raisebox{-4.3ex}[0pt]{\scriptsize Places} & \raisebox{-2.3ex}[0pt]{\scriptsize   $10M$ }  &
\raisebox{-1.3ex}[0pt]{ \scriptsize $434$ }&\raisebox{-1.3ex}[0pt]{ \scriptsize  $-$}&\raisebox{-1.3ex}[0pt]{ \scriptsize  $-$}
& \raisebox{-1.3ex}[0pt]{ \scriptsize $256\times256$}&\raisebox{-1.3ex}[0pt]{ \scriptsize  $2014$} & \scriptsize  The largest labeled dataset for scene recognition; Four subsets Places365 Standard, Places365 Challenge, Places 205 and Places88 as benchmarks. \\
\hline
\raisebox{-3.3ex}[0pt]{\scriptsize Open Images} & \raisebox{-3.3ex}[0pt]{\scriptsize   $9M$ } &\raisebox{-3.3ex}[0pt]{ \scriptsize $6000$+ }&\raisebox{-3.3ex}[0pt]{ \scriptsize  $-$}&\raisebox{-3.3ex}[0pt]{ \scriptsize  $8.3$}
& \raisebox{-3.3ex}[0pt]{ \scriptsize varied}&\raisebox{-3.3ex}[0pt]{ \scriptsize  $2017$} & \scriptsize Annotated with image level labels, object bounding boxes and visual relationships; Open Images V5 supports large scale object detection, object instance segmentation and visual relationship detection. \\
\hline
\end{tabular}
}
\end{table*}

\begin{table*}[!t]
\caption {Commonly used databases for metric learning.} \label{tab:metric_data}
\centering
\renewcommand{\arraystretch}{1.2}
\setlength\arrayrulewidth{0.2mm}
\setlength\tabcolsep{1pt}
\resizebox*{18cm}{!}{
\begin{tabular}{!{\vrule width1.2bp}c|c|c|c|c|c|c|p{8cm}!{\vrule width1.2bp}}
\hline
 \scriptsize   \shortstack [c] {\textbf{Dataset}\\ \textbf{Name}} & \scriptsize   \shortstack [c]
{\textbf{Total} \\ \textbf{Images}} & \scriptsize   \shortstack [c]
{\textbf{Categories}} & \scriptsize   \shortstack [c]
{\textbf{Training} \\ \textbf{Images}} & \scriptsize   \shortstack [c]
{\textbf{Testing }\\ \textbf{Images}} & \scriptsize   \shortstack [c] {\textbf{Retrieval} \\ \textbf{Level} }
& \scriptsize   \shortstack [c] {\textbf{Year}}
&\raisebox{1.3ex}[0pt]{ \scriptsize   \shortstack [c] {$\quad\quad\quad\quad\quad\quad\quad\quad\quad\quad\quad\quad$\textbf{Highlights}}} \\
\hline
\raisebox{-3.3ex}[0pt]{\scriptsize \shortstack [c] {CUB-200-2011} } &\raisebox{-3.3ex}[0pt]{\scriptsize   $11,788$ }& \raisebox{-3.3ex}[0pt]{\scriptsize $200$} & \raisebox{-3.3ex}[0pt]{ \scriptsize  $5,864$}& \raisebox{-3.3ex}[0pt]{\scriptsize $5,924$}
& \raisebox{-3.3ex}[0pt]{\scriptsize Object}& \raisebox{-3.3ex}[0pt]{ \scriptsize  $2011$} & \scriptsize  Fine-graind bird retrieval dataset \\
\hline
\raisebox{-3.3ex}[0pt]{\scriptsize CAR-196} & \raisebox{-3.3ex}[0pt]{\scriptsize   $16,185$ } &
\raisebox{-3.3ex}[0pt]{\scriptsize $196$} &
\raisebox{-3.3ex}[0pt]{ \scriptsize $8,054$} &\raisebox{-3.3ex}[0pt]{ \scriptsize  $8,131$}
&\raisebox{-3.3ex}[0pt]{ \scriptsize  Object}
&\raisebox{-3.3ex}[0pt]{ \scriptsize  $2013$} & \scriptsize  Fine-grained car retrieval dataset \\
\hline
\raisebox{-3.3ex}[0pt]{\scriptsize Standford Online} & \raisebox{-3.3ex}[0pt]{\scriptsize   $120,053$ } & \raisebox{-3.3ex}[0pt]{\scriptsize $22,634$} & \raisebox{-3.3ex}[0pt]{ \scriptsize  $59,551$}
 & \raisebox{-3.3ex}[0pt]{ \scriptsize  $60,502$}
& \raisebox{-3.3ex}[0pt]{ \scriptsize Instance}& \raisebox{-3.3ex}[0pt]{\scriptsize  $2016$} & \scriptsize Covers a variety of online shopping instances. \\
\hline
\raisebox{-3.0ex}[0pt]{\scriptsize Market-1501} & \raisebox{-3.0ex}[0pt]{\scriptsize  $32,000$  }  &
\raisebox{-3.0ex}[0pt]{ \scriptsize $1501$ }&\raisebox{-3.0ex}[0pt]{ \scriptsize  750 id}&\raisebox{-3.0ex}[0pt]{ \scriptsize  751 id}
& \raisebox{-3.0ex}[0pt]{ \scriptsize Instance}&\raisebox{-3.0ex}[0pt]{ \scriptsize  $2015$} & \scriptsize  One of the most widely used person re-identification dataset \\
\hline
\end{tabular}
}
\end{table*}

The chase of more and more annotated data in today's machine learning
community is a clear evidence of supervision's role in bridging the
semantic gap. A tremendous amount of efforts have been spent in data
collection and labeling nowadays. In the following, we will highlight
several datasets that are critical to the development of object
detection and metric learning techniques. 

\begin{figure*}[h]
\centering
\includegraphics[width=1.0\linewidth]{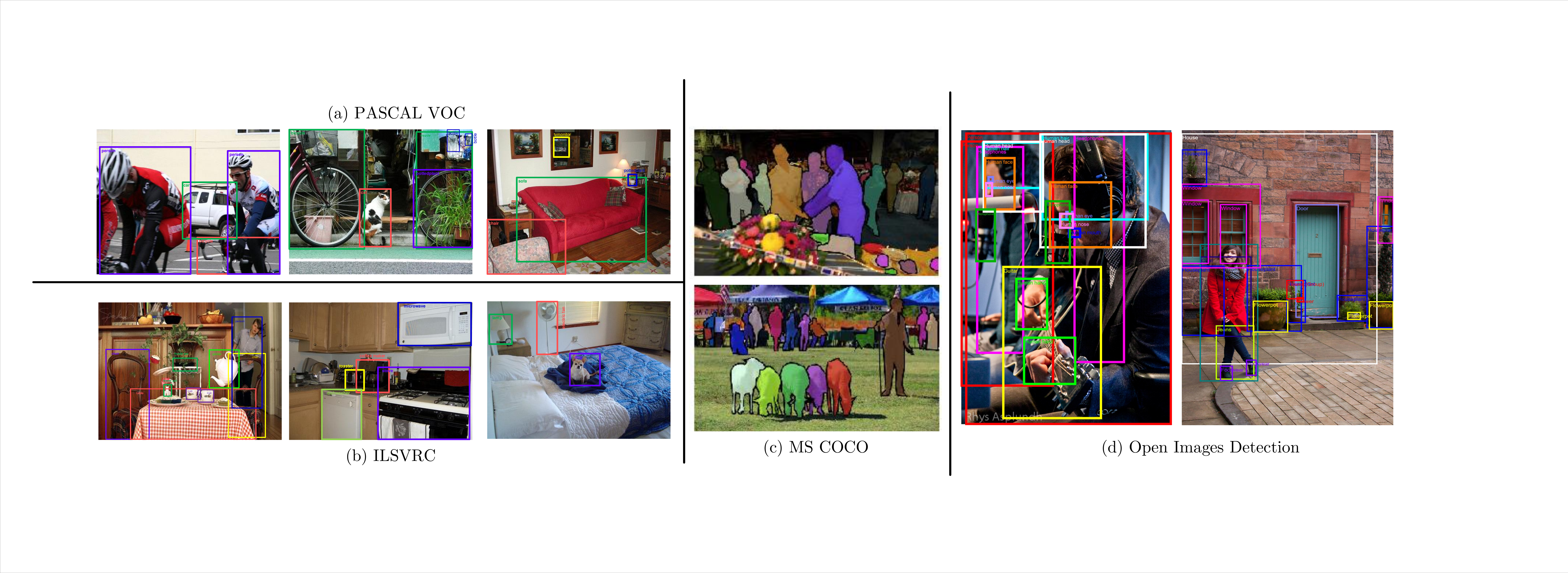}
\caption{Selected sample images for popular object detection datasets~\cite{everingham2010pascal,deng2009imagenet,lin2014microsoft,krasin2017openimages}}\label{fig:obj-data}
\end{figure*}

Four datasets are commonly used for generic object detection: PASCAL
VOC~\cite{everingham2010pascal}, ImageNet~\cite{deng2009imagenet}, MS
COCO~\cite{lin2014microsoft}, and Open
Imges~\cite{krasin2017openimages}.  The attributes of these datasets are
summarized in Table~\ref{tab:obj_data}.  The selected samples are shown
in Fig.~\ref{fig:obj-data}. Several criteria are used in evaluating the
performance of an object detector, including precision, recall, model
sizes, and inference speed measured by frames per second (FPS).  While
the average precision (AP) that combines precision and recall is used to
evaluate the performance for a specific category, the mean AP (mAP)
averaged over all categories is used as the measure of performance over
all categories.  For more details, readers are referred
to~\cite{lin2014microsoft}. 

\begin{figure*}[h]
\centering
\includegraphics[width=1.0\linewidth]{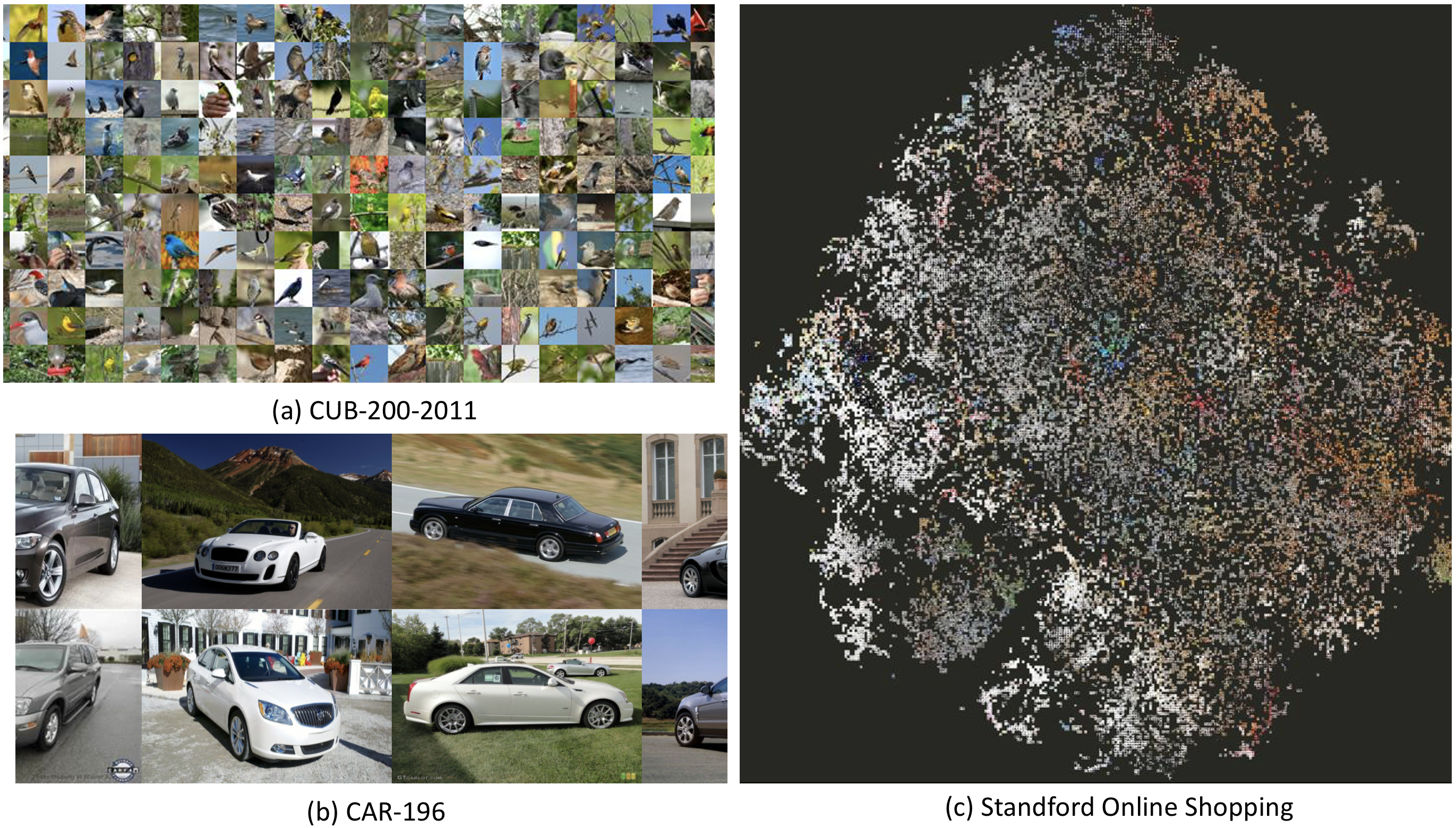}
\caption{Selected sample images for popular metric learning datasets\cite{wah2011caltech,krause20133d,oh2016deep,zheng2015scalable}. Subfigure (c) is a tSNE~\cite{van2008visualizing} visualization of the dataset~\cite{oh2016deep}.}
\label{fig:metric_data}
\end{figure*}

There are also four datasets that are commonly been used in metric
learning. They are: CUB-200-2011~\cite{wah2011caltech},
CAR-196~\cite{krause20133d}, Standford Online
Shopping~\cite{oh2016deep} and Market-1501~\cite{zheng2015scalable}. The
first two focus on fine-grained object category retrieval. The last two
are instance-level retrieval datasets. Market-1501 is one of the largest
person-reidentification benchmark dataset. Detailed statistics are shown
in Table~\ref{tab:metric_data} and exemplary images are shown in
Fig.~\ref{fig:metric_data}.  Precision$@k$, denoted by P$@k$, is a
popular metric in metric learning. It indicates the number of relevant
images among the top $k$ retrieved images. If there are $R$ images that
belong to the same class as the query, the R-precision (RP) measures the
percentage of correct retrievals among the top $R$ retrieved results.
Another recently proposed metric is MAP$@$R~\cite{musgrave2020metric},
that combines the idea of mean average precision with RP to offer a more
accurate performance measure. 

\section{Supervision for Object Detection}\label{sec:obj}

\subsection{Full Supervision}

Deep learning methods have been extensively developed for the
fully-supervised object detection task~\cite{girshick2014rich,
he2015spatial, girshick2015fast, sermanet2013overfeat,ren2015faster}.
Research on this topic has reached quite a mature stage. Generally
speaking, deep-learning-based object detection methods can be
categorized into two types: two-stage detection and one-stage detection.
Recently, there's an emerging line of transformer-based works~\cite{carion2020end,zhu2020deformable,dai2021up} which approach object detection as a direct set prediction problem. We elaborate representatives for each of the category below.

{\bf Two-Stage Detection.} Two-stage detection methods consist of two
stages in cascade: 1) the region proposal stage and 2) the object
classifiction stage. The common pipeline includes: category-independent
region proposals~\footnote{Object proposals,also called region proposals
or detection proposals, are a set of candidate regions or bounding boxes
in an image that may potentially contain an object} are generated from
an image, CNN features are extracted from these regions, and then
category-specific classifiers are used to determine the category label
of each proposal. 

\begin{figure*}[h]
\centering
\includegraphics[width=0.8\linewidth]{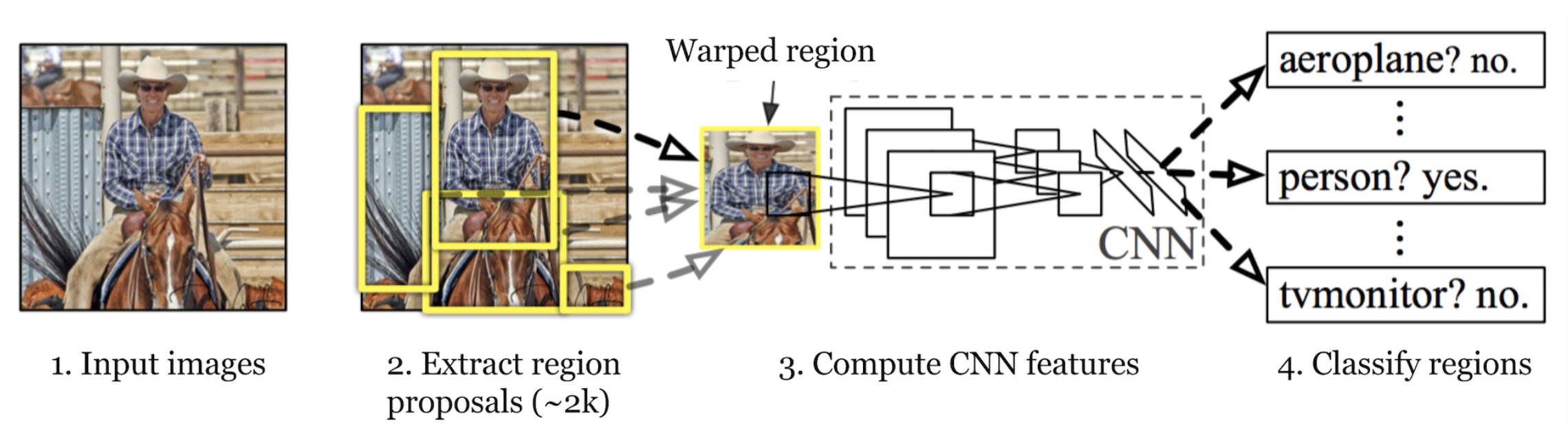}
\caption{The pipeline of a two-stage object detection framework
\cite{girshick2014rich}.}\label{fig:RCNN}
\end{figure*}

\begin{figure*}[h]
\centering
\includegraphics[width=0.7\linewidth]{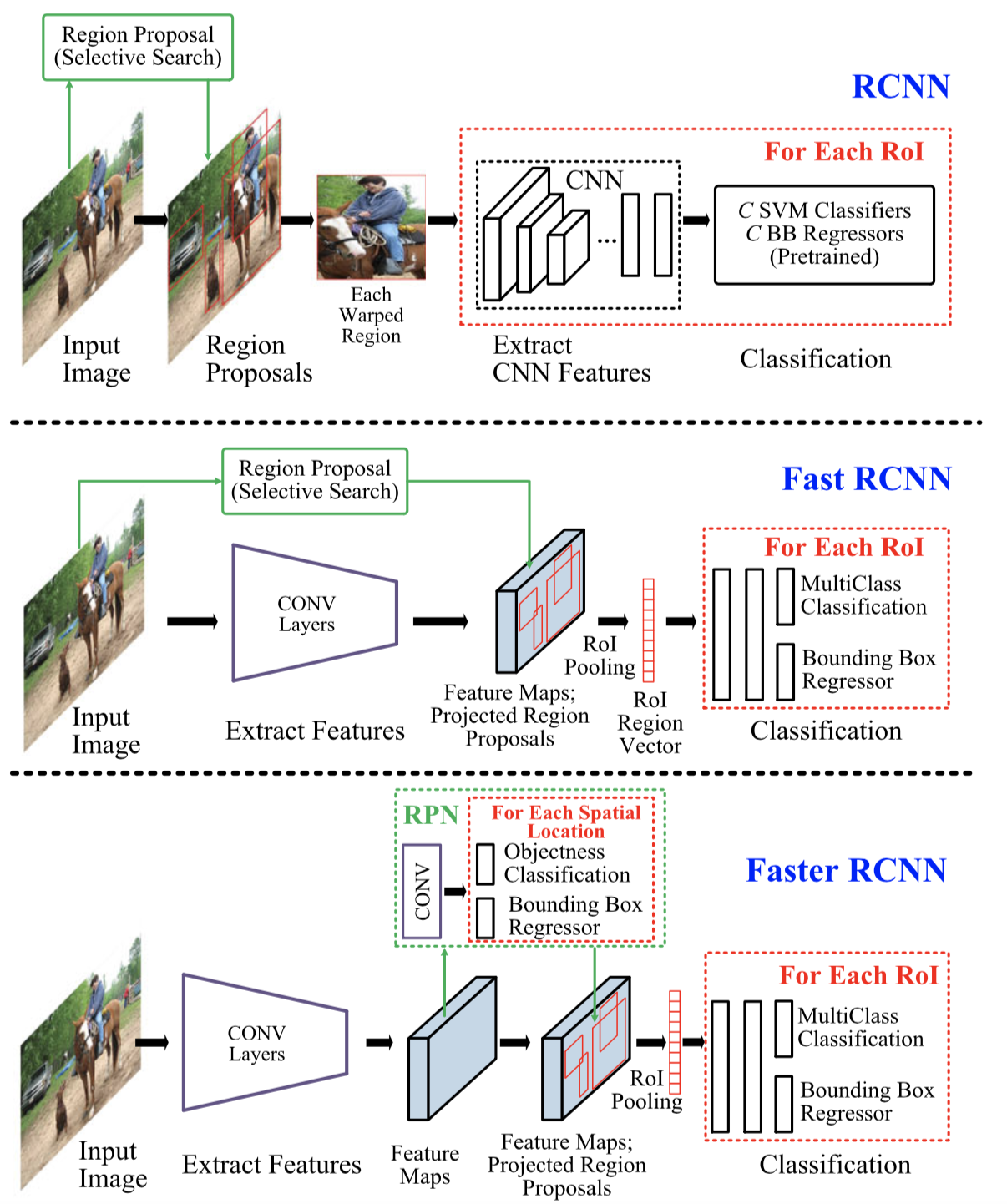}
\caption{The system diagrams of three two-stage object detection
methods~\cite{liu2020deep}: RCNN (top), Fast RCNN (middle), and Faster RCNN (bottom).}
\label{fig:Fast-RCNN}
\end{figure*}

\begin{figure*}[h]
\centering
\includegraphics[width=0.7\linewidth]{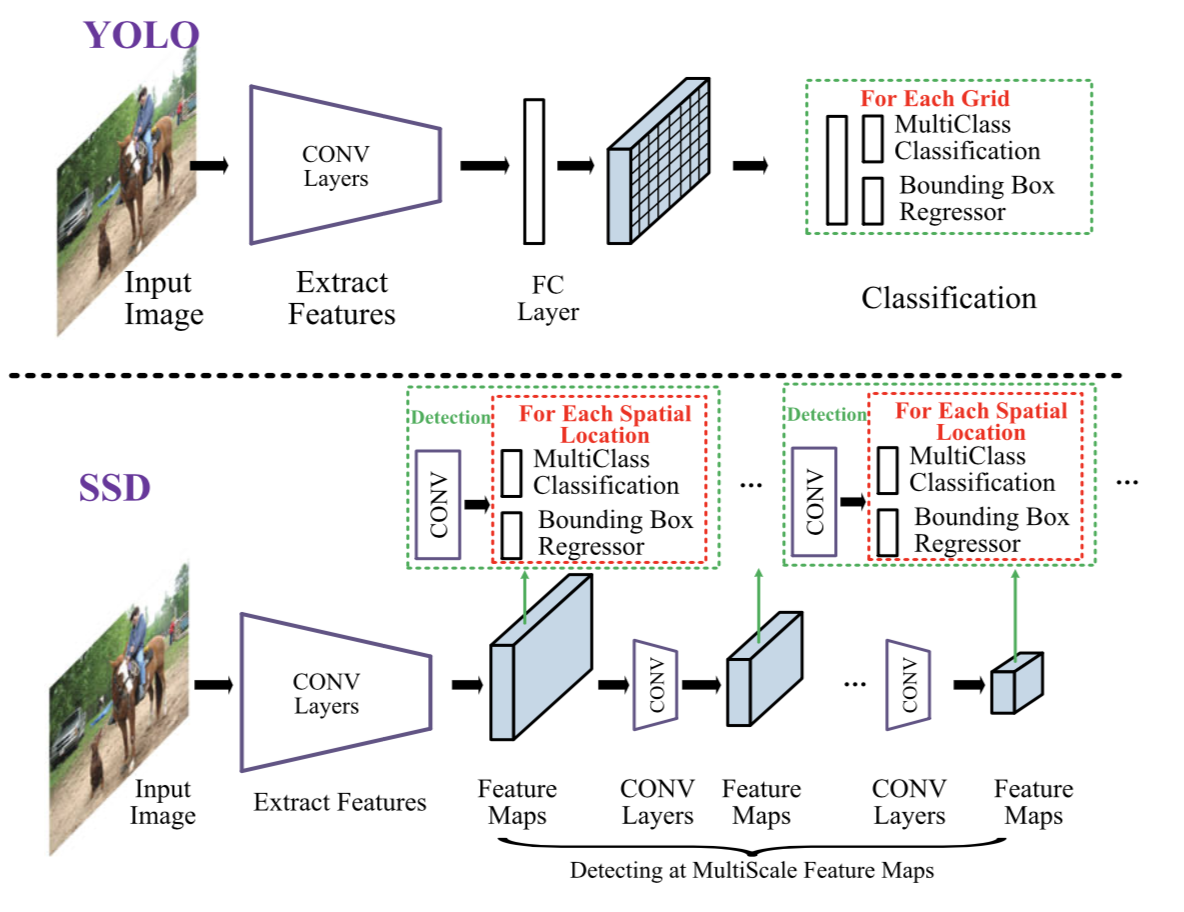}
\caption{The system diagrams of two one-stage object detection
networks~\cite{liu2020deep}: YOLO (top) and SDD (bottom).}\label{fig:SSD}
\end{figure*}

Inspired by the impressive image classification performance achieved by
the AlexNet \cite{krizhevsky2012imagenet} and the success of selective
search in finding region proposals with hand-crafted
features~\cite{uijlings2013selective}, Girshick et
al.~\cite{girshick2014rich,girshick2015region} were among the first to
explore CNNs for generic object detection and developed RCNN as shown in
Fig.~\ref{fig:RCNN}. The training of an RCNN framework consists of the
following steps.
\begin{enumerate}
\item Region proposal selection \\
Class agnostic region proposals, which are candidate regions that might
contain objects, are obtained via selective search. 
\item Region proposal processing \\
Selected region proposals are cropped from the image and warped into the same size. They are used as the input to finetune a CNN model
pre-trained by a large-scale dataset such as ImageNet. In this step, a
region proposal with its IOU against a ground truth box greater than
0.5~\footnote{This is a commonly used practice such as in MSCOCO
dataset~\cite{lin2014microsoft}.} is defined as a positive for the
ground truth class and the rest as negatives. 
\item Class-specific SVM classifiers training \\
A set of class-specific linear SVM classifiers are trained using the
fixed length features extracted by the CNN, replacing the softmax
classifier learned by finetuning. For the training of SVM classifiers,
positive examples are the ground truth boxes for each class. A region
proposal that has less than 0.3 IOU with all ground truth instances of a
class is negative for that class. Note that the positive and negative
examples used for training SVM classifiers are different from those for
finetuning the CNN. 
\item Class-specific bounding box regressor training \\
Bounding box regression is learned for each object class with CNN
features. 
\end{enumerate}

There are variants of RCNN for better performancne. Two noticeable ones
are Fast RCNN \cite{girshick2015fast} and Faster
RCNN~\cite{ren2015faster} as shown in Fig. \ref{fig:Fast-RCNN}.  Fast
RCNN improves both detection speed and accuracy of RCNN.  Rather than
separately training a softmax classifier, SVMs, and bounding box
regressors as done in RCNN, Fast RCNN enables end-to-end detector
training by developing a streamlined training process that
simultaneously learns a softmax classifier and class-specific bounding
box regression. The core idea of Fast RCNN is to share the feature
extraction process among different region proposals.  Fast RCNN improves
efficiency of RCNN considerably, typically 3 times faster in training
and 10 times faster in testing, and there is no storage required for
feature caching.  Faster RCNN offers an efficient and accurate region
proposal network (RPN) in generating region proposals. It utilizes the
same backbone network but exploits features from the last shared
convolutional layer to accomplish the task of RPN for region proposal
generation and the task of Fast RCNN for region classification. 

The two-stage region-based pipeline offers state-of-the-art object
detection performance as evidenced by the fact that leading results on
popular benchmark datasets are all based on Faster
RCNN~\cite{ren2015faster}. Nevertheless, region-based methods are
computationally costly for mobile/wearable devices with limited storage
and computational resources. Instead of optimizing individual components
of the complex region-based pipeline, researchers looked for an
alternative that detects objects directly without the region proposal
step. 

{\bf One-Stage Detection.} By one-stage detection, we refer to an
architecture that predicts class probabilities and bounding box sizes
and locations from full images with a single feed-forward CNN in a
monolithic setting. It can be optimized end-to-end directly on detection
performance.  DetectorNet~\cite{szegedy2013deep} was among the first in
exploring this new direction. However, since the network needs to be
trained per object type and mask type, it does not scale well as the
number of classes increases.  Redmon et al.~\cite{redmon2016you}
proposed YOLO (You Only Look Once), which is a unified detector casting
object detection as a regression problem from image pixels to spatially
separated bounding boxes and associated class probabilities.  Unlike the
two-stage detection, YOLO predicts detections based on features from
local regions of multiple sizes. Specifically, YOLO divides an image
into an $S \times S$ grid, each predicting $C$ class probabilities, $B$
bounding box locations, and confidence scores. By eliminating the region
proposal generation step entirely, YOLO is fast by design, running in
real time at 45 FPS. Fast YOLO~\cite{shafiee2017fast} can even reach 155
FPS.  To preserve real-time speed without sacrificing much detection
accuracy, Liu et al.~\cite{liu2016ssd} proposed SSD (Single Shot
Detector), which is faster than YOLO and with an accuracy competitive
with region-based detectors such as Faster RCNN~\cite{ren2015faster}.
SSD effectively combines ideas from RPN in Faster RCNN and YOLO
multiscale CONV features to achieve fast detection speed, while still
retaining high detection quality. Like YOLO, SSD predicts a fixed number
of bounding boxes and scores, followed by a non-maximum-suppression
(NMS) step to produce the final detection. The system diagrams of YOLO
and SSD are shown in Fig.~\ref{fig:SSD} for comparison.


Recently, there's a growing trend in applying transformers to the computer vision tasks~\cite{dosovitskiy2020image,touvron2021training,carion2020end}, among which DETR~\cite{carion2020end} is a representative for object detection. Instead of generating ``proposals'', it patchfies the given image as tokens and feeds them to the vision transformer. While previous detectors rely on NMS as a post-processing step, DETR casts object detection as a set prediction problem and leverages the Hungarian algorithm to match the box predictions with ground-truth boxes during training. DETR simplifies traditional object detection pipeline and obtains 42 AP on COCO using a Resnet50 backbone.

\begin{figure}[h]
\centering
\includegraphics[width=1.0\linewidth]{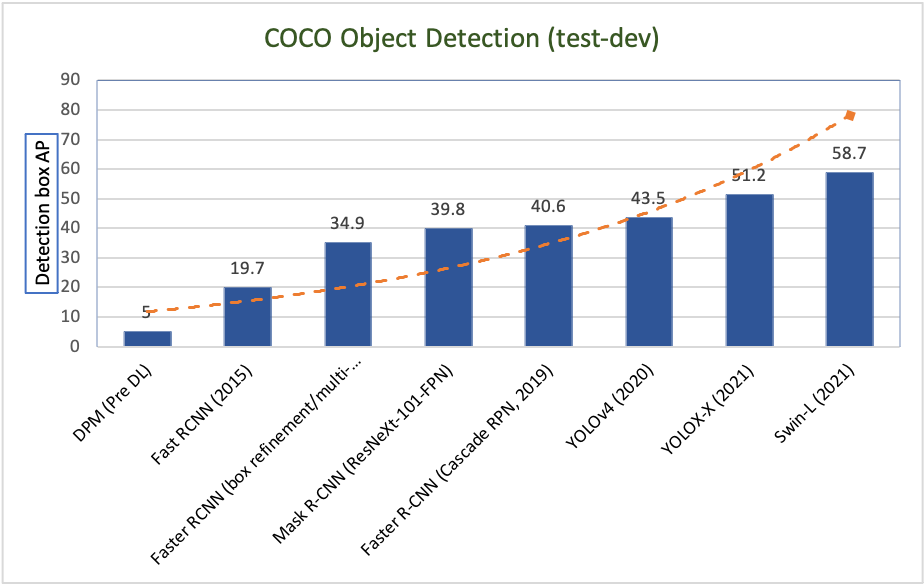}
\caption{Progress of object detection performance on Microsoft COCO over years, where results are quoted from~\cite{pwc2021coco}.} \label{fig:COCO}
\end{figure}

{\bf Further Performance Improvement.} We show the detection
performance improvement over years with respect to the Microsoft COCO
challenge in Fig.~\ref{fig:COCO}. As to the object detection task in the
open image challenge, the current leader~\cite{liu2021swin} achieved 58.7 box AP in the public leader board. It  proposes a hierarchical transformer whose
representation is computed with Shifted windows. Representative methods benchmarked include 
Fast RCNN~\cite{girshick2015fast}, Faster-RCNN~\cite{ren2015faster},
FPN~\cite{ghiasi2019fpn}, Deformable Faster
RCNN~\cite{ren2018deformable}, Cascade RCNN~\cite{cai2018cascade}, Mask-RCNN~\cite{he2017mask} and YOLO families~\cite{redmon2016you}.
Generally speaking, the backbone network, the detection pipeline and the
availability of large-scale training datasets are three most important
factors in further detection accuracy improvement. Besides, ensembles of
multiple models, the incorporation of context features, and data
augmentation all help achieve better accuracy. 

\subsection{Weak Supervision}

Object detectors are trained without bounding box annotations in weak
supervision detection (WSD), where only image-level labels are used. The
main challenge of WSD is object localization since a label may refer to
any object in the image. This problem is typically addressed using
multiple instance learning, which is a well-studied topic
~\cite{bilen2016weakly,wan2019c,cinbis2016weakly}. Although image-level
labels are easier to collect than bounding boxes, they still require
manual efforts. Besides, they are often limited to a pre-defined
taxonomy. 

Some recent work adopts captions, which are often freely available on
the web. Learning object detection from captions has been studied but at
a limited scale.  CAP2Det~\cite{ye2019cap2det} parses captions into a
multi-label classification targets and, then, these labels are used to
train a WSD model. However, it requires image-level labels to train the
caption parsers. Besides, it is under the constraint of a closed
vocabulary.  Another WSD model was trained in \cite{amrani2019learning}
based on a predefined set of words in captions. It is similar to a
closed vocabulary, yet the rich semantic content of captions is
discarded.  In contrast, research in~\cite{sun2015automatic} and
~\cite{ye2018learning} aims to discover an open set of object classes
from image-caption corpora, and learns detectors for each discovered
class. 

One shortcoming of WSD methods is their poorer object localization
accuracy. Object recognition and localization are disentangled into two
independent problems in \cite{zareian2021open}. It learns object
localization using a fully annotated dataset from a small subset of
classes and conducts object detection using open-vocabulary captions. 

\subsection{Mixed Full/Weak Supervision}

Mixed supervision has been studied to exploit both full supervision and
weak supervision. Most mixed supervision methods need bounding box
annotations for all main classes and apply weak supervision to auxiliary
classes~\cite{gao2019note,ramanathan2020dlwl,wang2015model}. For
example, by following the transfer learning framework, one can transfer
a detector trained on supervised base classes to weakly supervised
target classes~\cite{hoffman2014lsda,tang2016large,
uijlings2018revisiting}.  These methods usually lose performance on
target classes.

One common limitation of mixed-supervised methods is that they require
image-level annotations within a predefined taxonomy so that they learn
the predefined classes only. To address it, one recent
work~\cite{desai2021virtex} exploits supervision from captions that are
open-vocabulary and more prevalent on the web. Instead of training on
base classes and transferring to target classes, it uses captions to
learn an open-vocabulary semantic space that includes target classes,
and transfers that to the object detection task via supervised learning. 


\subsection{Zero-shot Detection}

Zero-shot object detection (ZSD) aims to generalize from seen object
classes to unseen ones by exploiting zero-shot learning techniques (e.g.
word embedding projection~\cite{frome2013devise}) for object proposal
classification.  There exist however different views on ZSD.  According
to~\cite{bansal2018zero}, the main challenge of ZSD lies in modelling
the background class since it is difficult to separate from unseen
classes. The background was treated as a mixture model in
~\cite{bansal2018zero}.  It was furthermore improved by introducing the
polarity loss~\cite{rahman2020improved}.  On the other hand, it was
argued in~\cite{zhu2020don} that the key challenge of ZSD is the
generalization capability of object proposal models. To tackel with it,
they employed a generative model to hallucinate unseen classes and
augment seen examples in the proposal model training process. 

Nevertheless, there is still a significant gap in the performance due to
the unnecessarily harsh constraint; namely, not having any example of
unseen objects, and having to guess how they look like solely based on
their word embeddings~\cite{bansal2018zero, rahman2020improved} or
textual descriptions~\cite{li2019zero}. This has motivated researchers
to simplify the task by making more assumptions such as the availability
of test data during training~\cite{rahman2019transductive} or the
availability of unseen class annotations to filter images with unseen
object instances~\cite{gupta2020multi}. Since datasets with natural,
weak supervision are abundant and cheap, an alternative was proposed to
utilize image-caption datasets in~\cite{zareian2021open}, which covers a
larger variety of objects with an open vocabulary. 

\section{Supervision for Metric Learning}\label{sec:cbir}

\subsection{Full Supervision}

Metric learning attempts to map image data to an embedding space, where
images of similar semantic content are closer while those of dissimilar
semantic meaning are farther apart. The embedding learned in this way captures semantics intuitive to human understanding which are initially not obvious in the pixel form. In general, this objective can be
achieved by leveraging embedding and/or classification losses. 

\begin{figure*}[h]
\centering
\includegraphics[width=0.7\linewidth]{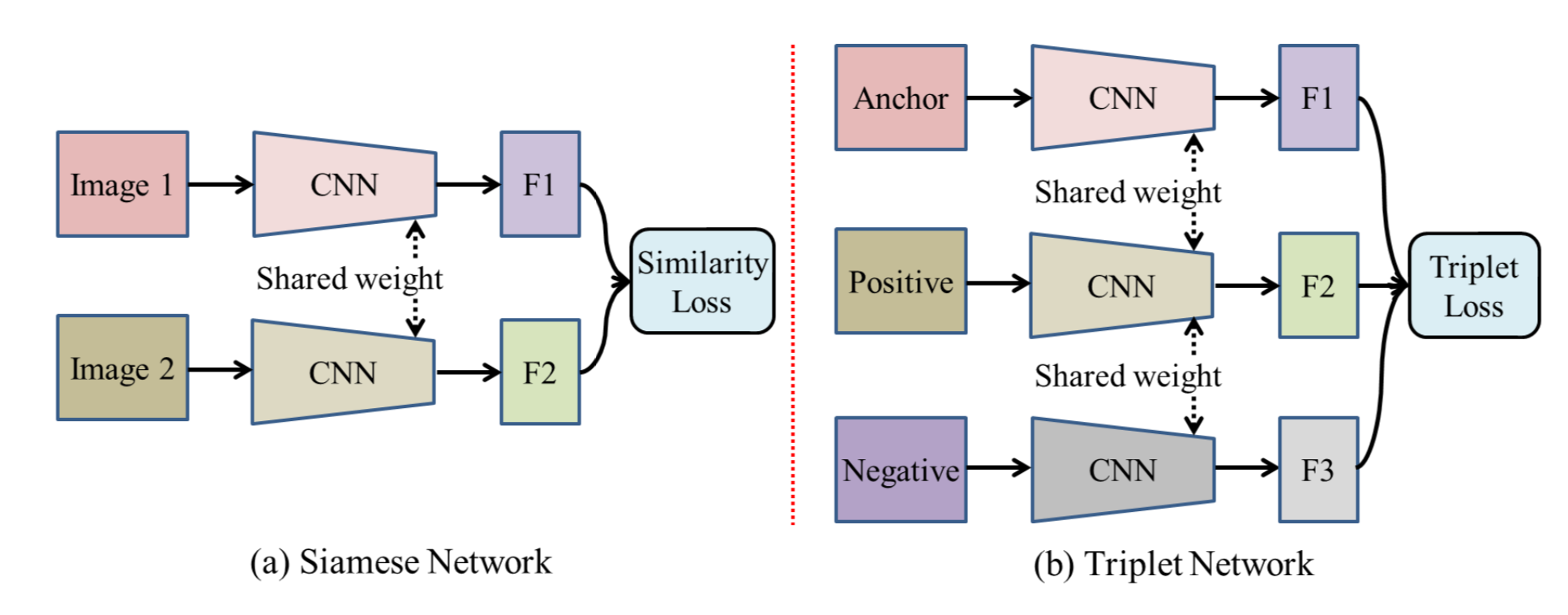}
\caption{Comparison of the similarity loss and the triplet loss 
using the siamese network~\cite{dubey2021decade}. For Siamese Network, it optimizes to increase the similarity between positive pairs and decrease the similarity between negative pairs. The Triplet Network enforces the distance between the anchor and the positive to be smaller than that between the anchor and negative.}\label{fig:siamese}
\end{figure*}

{\bf Embedding Losses.} The embedding loss operates on the relationship
between samples in a batch while the classification losses include a
weight matrix that transforms the embedding space into a vector of class
logits.  Typically, embeddings are preferred when the task is in form of
information retrieval whose goal is to return a data sample that is most
similar to the query one. A specific example is image retrieval, where
the input is a query image and the output is the most similar image in a
database.  Another application context is open-set classification where
the test set and the training set classes are disjoint, and there are
cases no proper classification loss can be defined.  For example, when
constructing a dataset, it might be difficult (or costly) to assign the
class label to each sample. It might be easier to specify the relative
similarities between samples in form of pair- or triplet-relationship.
Pairs and triplets can provide additional information for existing
datasets. Since both do not have explicit labels, embedding losses
become a suitable choice.  Pair and triplet losses provide the
foundation to two fundamental metric learning approaches (See Fig.~\ref{fig:siamese}). 

{\bf Contrastive Loss.} A classic pair-based loss is the contrastive
loss~\cite{hadsell2006dimensionality} in form of
$$
L_{\rm contrastive} = [d_p - m_{pos}]_{+} + [m_{neg} - d_n]_{+},
$$
where notation $[x]_{+}$ denotes $\max(x,0)$. In the above equation, it attempts to make the distance between positive pairs $d_p$ smaller
than threshold $m_{pos}$, and the distance between negative pairs $d_n$
larger than threshold $m_{neg}$. A theoretical downside of this method is
that the same distance threshold is applied to all pairs even though
there may be a large variance in their similarities and dissimilarities.
The triplet margin loss~\cite{weinberger2009distance} is developed to
address this issue. 

{\bf Triplet Loss.} A triplet consists of an anchor input, $A$, a
positive input, $P$, and a negative input $N$, where the anchor is more
similar to the positive than the negative. The triplet margin loss is
used to ensure that the anchor-positive distance ($d_{ap}$) is smaller
than the anchor-negative distance ($d_{an}$) by a predefined margin
($m$). The triplet loss function can be written in form of
$$
L_{\rm triple}=\left[|| f(A)-f (P)||^2 - || f(A)-f (N)||^2 + m 
\right]_{+},
$$
where $f$ is an embedding function.  This triplet loss places fewer
restrictions than the contrast loss in the embedding space.  It allows
a learned model to account for the variance in interclass dissimilarities. 

{\bf Other Loss Functions.} A wide variety of loss functions has been
defined based on these fundamental concepts. For example, the angular
loss~\cite{wang2017deep} is a triplet loss where the margin is based on
the angles formed by the triplet vectors. The margin
loss~\cite{wu2017sampling} modifies the contrastive loss by setting $$
m_{pos}=\beta - \alpha, \mbox{ and } m_{neg}=\beta + \alpha, $$ where
$\alpha$ is fixed, and $\beta$ is learnable. Other pair losses are based
on the softmax function and LogSumExp, which is a smooth approximation
of the maximum function. Specifically, the lifted structure
loss~\cite{oh2016deep} is the contrastive loss but with Log-SumExp
applied to all negative pairs.  The N-Pairs loss~\cite{sohn2016improved}
applies the softmax function to each positive pair relative to all other
pairs. It is also known as InfoNCE~\cite{oord2018representation} and
NT-Xent~\cite{chen2020simple}.  The tuplet margin loss~\cite{yu2019deep}
combines Log-SumExp with an implicit pair weighting method while the
circle loss~\cite{sun2020circle} weighs each pair's similarity by its
deviation from a pre-determined optimal similarity value.  A general
weighting framework was presented in \cite{wang2019multi} to understand
recent pair-based loss functions.  In contrast with pair and triplet
losses, FastAP~\cite{cakir2019deep} attempts to optimize for average
precision within each batch using a soft histogram binning technique. 

{\bf Classification Losses.} Classification losses are obtained by
including of a weight matrix, where each column corresponds to a
particular class. In most cases, the training process consists of
multiplying weight matrix with embedding vectors to obtain logits, and
then applying a certain loss function to the logits. The most
straightforward one is the normalized softmax
loss~\cite{wang2017normface, liu2017sphereface, zhai2018classification}.
It is identical with the cross entropy loss with L2-normalized columns of
the weight matrix.  

One variant is ProxyNCA~\cite{movshovitz2017no}, where the cross entropy
loss is applied to the Euclidean distances, rather than the cosine
similarities, between embeddings and the weight matrix. A number of face
verification losses modified the cross entropy loss with angular margins
in the softmax expression. For example,
SphereFace~\cite{liu2017sphereface}, CosFace~\cite{wang2018additive,
wang2018cosface} and ArcFace~\cite{deng2019arcface} apply
multiplicative-angular, additive-cosine and additive-angular margins,
respectively. It is interesting to note that many metric learning papers
leave out face verification losses from their experiments although they
are not face-specific. The SoftTriple loss~\cite{qian2019softtriple}
takes a different approach by expanding the weight matrix to have
multiple columns per class. It has more flexibility in modeling class
variances. 

\subsection{Other Forms of Supervision}
Embedding of semantic information is hard to learn when the amount of labeled data is limited or when the data is imbalanced, which is often the case in real-world scenario. The research community tries to address the semantic issues by either adopting the weakly labeled data, partially labeled data or unlabeled data, leading to different supervisions discussed below.

{\bf Weak Supervision.} Weakly supervised approaches have been explored
for the image retrieval task \cite{gattupalli2019weakly, guan2018tag,
li2020weakly, tang2017weakly}.  For example, Tang et al.
\cite{tang2017weakly} proposed a weakly-supervised multimodal hashing
method that exploits local discriminative and geometric structures in
the visual space.  Guan et al.\cite{guan2018tag} performed pre-training
in weak supervision mode and finetuned the network in supervision mode.
Gattupalli et al.\cite{gattupalli2019weakly} developed a weakly
supervised deep hashing method that used tag embeddings for image
retrieval with the word2vec semantic embeddings. Li et
al.\cite{li2020weakly} developed a semantic guided hashing network for
image retrieval by employing the weakly-supervised tag information and
inherent data relations simultaneously. 

{\bf Semi-Supervision.} The semi-supervised approaches generally use a
combination of labeled and unlabelled data in feature learning.  A
semi-supervised deep hashing framework was proposed for image retrieval
from labeled and unlabeled data in \cite{zhang2017ssdh}. It uses labeled
data for empirical error minimization and both labeled and unlabeled
data for embedding error minimization.  The generative adversarial
learning approach was also utilized in semi-supervised deep image
retrieval \cite{jin2020ssah,wang2018semi}. A teacher-student
semi-supervised image retrieval method was presented in
\cite{zhang2019pairwise}, where the pairwise information learned by the
teacher network is used as the guidance to train the student network. 
Pseudo labels are another source of supervision falling into semi-supervised regime.  For example, \cite{hu2017pseudo} generates pseudo labels based on the pretrained VGG16 features via k-means clustering. In~\cite{duan2021slade} a self-training framework, SLADE, is proposed to improve retrieval performance by leveraging additional unlabeled data. It first train a teacher model on the labeled data and use it to generate pseudo labels for the unlabeled data. It then train a student model on both labels and pseudo labels to generate final feature embeddings. The framework significantly improves existing state-of-the-art.

{\bf Self Supervision.} There has been an arising amount of interest in
self-supervised learning~\cite{he2020momentum,chen2020simple}. It is similar to unsupervised learning but
apply self-generated pseudo-labels to the data during the training
process. Self supervision is often achieved by clever usage of data
augmentation or information from other modalities. For example, \cite{misra2020self} defines a set of pretext tasks for learning invariant feature representation. In \cite{caron2021emerging}, an online vision transformer was asked to predict the output of a target vision transformer, whose input is an augmentation of the first transformer's input. As this requires no annotations, it is self-supervised and exhibits superior performance when applied to image retrieval. Self-supervised learning also proves useful for initializing deep metric learning embedding~\cite{duan2021slade}, video retrieval~\cite{zhang2016play}, and cross-image retrieval~\cite{li2018self}.

{\bf No Supervision.} Though supervised models have shown promising
performance in image retrieval, it is always difficult to get labeled
large-scale data.  Thus, unsupervised models have been investigated and
they do not require class labels to learn features. Generally,
unsupervised models enforce the constraints on hash codes and/or
generate the output to learn features.  Erin et al.~\cite{erin2015deep}
used deep networks in an unsupervised manner to learn hash codes with
the help of constraints such as the quantization loss, balanced bits and
independent bits.  Huang et al.~\cite{huang2016unsupervised} utilized
deep networks coupled with unsupervised discriminative clustering to
learn the description in an unsupervised manner. Paulin et al.
\cite{paulin2015local} used an unsupervised convolutional kernel network
to learn convolutional features for image retrieval.  They applied it to
patch retrieval as well.  

Lin et al.~\cite{lin2016learning} imposed constraints (e.g., the minimal
quantization loss, evenly distributed codes, and uncorrelated bits) on
an unsupervised deep network and proposed a solution, called DeepBit,
for image retrieval, image matching and object recognition applications.
DeepBit has a two-stage training process.  In the first stage, the model
is trained with respect to above-mentioned objectives. To improve its
robustness, the network is finetuned in the second stage based on
rotation data augmentation. The analysis of DeepBit is given in
\cite{lin2018unsupervised}.  However, DeepBit suffers from severe
quantization loss due to rigid binarization of data using the sign
function without considering its distribution property. To tackle the
quantization problem of DeepBit, a deep binary descriptor with
multiquantization was proposed by Duan et al.  \cite{duan2017learning}.
It is achieved by jointly learning the parameters and the binarization
functions using a K-AutoEncoders (KAEs) network. 

\section{Future Research Directions}\label{sec:future}

Supervision has been dominated in form of data labeling in the last
decade. However, this form appears to be quite limited. Humans learn
semantics and knowledge from a wide range of resources, e.g., domain
knowledge priors, correlation from different domains and modalities,
etc.  Although it is still a mystery how humans learn semantic meanings
from the real world, it is anticipated that supervision on machines will
appear in richer form. In this section, we present two general
directions for future research. 

\subsection{Interpretable and Modularized Learning}\label{subsec:SSL}

Interpretability and modular design are pillars to the construction,
debugging and maintenance of next-generation artificial intelligence
(AI) systems.  Although deep learning is the dominant methodology in
providing the mapping between image pixels and semantics nowadays, it is
neither interpretable nor modularized and we anticipate the same mapping
to be achieved by other alternatives. 

One emerging alternative is successive subspace learning
\cite{chen2018saak, kuo2016understanding, kuo2017cnn, kuo2018data,
kuo2019interpretable, rouhsedaghat2021successive}.  Simply speaking, SSL
is a light-weight unsupervised data embedding (or feature learning)
method and it can be applied to different data types (e.g. images,
point-clouds, voxels, etc.) The SSL pipeline consists of a sequence of
joint spatial-spectral transforms in cascade with PCA-like transform
kernels.  They are rigorously derived using statistical properties of
data units such as pixels, voxels and points. SSL-based embedding is
data driven and repeatable. The SSL pipeline can be connected to a
classifier (e.g., the random forest, the support vector machine or the
extreme gradient boosting classifier, etc.) or a regressor (e.g., the
linear regressor, the logistic regressor, the support vector regressor,
etc.) for final decision. 

The representations associated with SSL are unsupervised, interpretable,
modularized, robust to perturbations, effective (i.e., a small embedded
dimension) and efficient (a smaller model size and low embedding
complexity). Since end-to-end optimization is completely abandoned in
SSL, its training complexity is significantly lower. It can be
implemented on low-cost CPUs. The sizes of SSL models are significantly
smaller than those of DL-based models; thus, suitable for mobile and
edge computing.  From the angle of supervision, SSL can incorporate
priors conveniently, and fine-tune the AI system with new observations
on the fly. 

SSL-based solutions find applications in object classification
\cite{chen2020pixelhop, chen2020pixelhop++, manimaran2020visualization,
yang2021pixelhop}, fake face image detection \cite{chen2021defakehop},
face gender classification \cite{rouhsedaghat2020facehop},
low-resolution face recognition \cite{rouhsedaghat2021low}, joint
compression and classification \cite{tseng2020interpretable}, point
cloud classification and registration \cite{kadam2020unsupervised,
kadam2021r, zhang2020pointhop++, zhang2020pointhop,
zhang2020unsupervised}, image and texture generation \cite{lei2020nites,
lei2021tghop}, anomaly detection \cite{zhang2021anomalyhop} and medical
image classification \cite{liu2021voxelhop}.

\subsection{Intelligence Gaps}
Intelligence gaps is a collection of three
characterized aspects to target, in order to reach human-level semantic
understanding from raw data perception. The three aspects envision three
increasing levels of semantic understanding which we examine how to fill in below. 

{\bf Gap between Signals and Semantic Units.} Humans have sensors such
as eyes, ears, nose, skin, etc. to receive signals (or stimuli) from the
external world.  They include visual, audio, smell, pressure,
temperature, etc. Signals need to be converted to compact
representations for future processing in machines - known as
``embedding''. We have witnessed rapid progress in signal/data
embedding.  There are a few criteria in evaluating embedding schemes:
interpretability, supervision degree, sensitivity, effectiveness and
efficiency. Most today's embedding methods rely on deep learning.  They
are far from ideal according to these criteria. A new signal embedding
idea is to exploit statistical properties of data units (e.g., pixels,
vertices, and points) in an unsupervised feedforward fashion based on
SSL as discussed in Sec.  \ref{subsec:SSL}. It is interpretable, robust
to perturbations, effective (i.e., smaller embedded dimension),
efficient (smaller model size and lower complexity), and suitable for
multi-tasking. 

Furthermore, two challenges in signal embedding worth further study: 1)
attention and 2) multi-modal data representation. Both machines and
human brains have limitations on processing speed, memory and
communication capacity. Attention is needed to enable an intelligent
system to process the most relevant input within its limits. Attention
is often derived by end-to-end optimization nowadays (e.g. visual
saliency in computer vision and transformers in natural language
processing). Yet, attention can be easily fooled with small
perturbation. For example, it can be shifted from one region to another
in an image by manipulating a few pixels - leading to a totally
different outcome.  Adversarial attacks impose a major threat in
real-world applications. Interpretable and robust attention is essential
in next generation AI, which will be assisted by semantic scene and
object segmentation. For multi-modal data representation, subspace
decomposition may be leveraged. That is, we may represent audio, image,
video, and 3D data in their individual subspaces and select a suitable
combination and use the direct sum of these subspaces to construct a
multi-modal space. Each subspace can be updated independently and
combined efficiently and dynamically in response to different needs. 

{\bf Gap between Semantic Units and Knowledge}. Semantic units are
segmented for ease of re-composition. The study of their relationship
yields a richer information space. For example, the WordNet contains
relations between numerous words so as to result in a huge graph.
Knowledge presents the highest abstraction level of human cognition.
Besides knowledge representation and acquisition, humans can infer
missing information and discover knowledge that are not directly
available. 

Generally, one can construct individual knowledge graphs based on
existing databases in various domains and then combine them into larger
heterogeneous graphs.  Knowledge graphs will be a central piece of the
next generation AI.  There are open problems to be addressed, including
scalability, ambiguity resolution, semantic matching, path
finding/completion, new entity discovery, hidden relation extraction,
dynamic graph evolution, etc. 

To tackle with scalability, a decomposition and re-composition
methodology through interaction of semantic and knowledge spaces could
be a direction to explore. For example, today's CNNs recognize cars of
different colors and models from various angles through numerous labeled
car images. Yet, this is not how humans acquire the knowledge of "cars".
Humans decompose cars into semantic units such as body, wheels, doors,
windows, lights, etc. and use them to form the knowledge of "cars". The
decomposition/re-composition process enables humans to learn cars with
fewer examples. The success lies in the interaction of the 3D car
structure (knowledge) and the projected 2D car images (semantic units).
Also, it is challenging to recognize small components of cars such as
wheels, doors, windows, lights, etc. alone.  Yet, the 3D structural
knowledge of cars can help trace/confirm the parts and make their
recognition easier. The same principle applies to human perception on
objects with occlusion. We can recognize occluded objects if occlusion
is not severe. Generally, one recognizes objects through their salient
regions and, then, their parts through the assistance of knowledge
graphs. 

Ambiguity resolution can be done using the context information in the
knowledge space. Ontology is essential to human knowledge acquisition,
organization, and learning.  Hierarchical categorization is more stable
and easier to update. Today's knowledge graphs are flat without ontology
incorporated. Lack of knowledge hierarchy makes the representation
difficult to scale up. It may be feasible to enforce the ontological
relationship in subspace decomposition. That is, a high-dimensional
knowledge space will be decomposed into a direct sum of multiple
low-dimension knowledge subspaces. The core knowledge, which is stored
in a low-dimension subspace, should be more stable and error resilient
with less frequent update. The refined knowledge in specific domains is
stored in other low-dimensional subspaces. They will be updated more
frequently and optimized locally. Unequal security can be applied to
protect different subspaces from attacks depending on their importance.
Knowledge space decomposition and re-composition provides flexibility in
face of a rapidly changing, stochastic and adversarial environment.
Mathematically, this decomposition can be achieved by tensor operations. 

{\bf Gap between Knowledge and Concept/Decision}. Knowledge is what we know. It's the accumulation of past experience and insight that shapes the lens by which we interpret, and assign meaning to, information. In psychology, decision-making is regarded as the cognitive process resulting in the selection of a belief or a course of action among several possible alternative options. Logical inference
is the basis of human reasoning. Although this is analogous to path
finding in knowledge graph, path finding is often used to find the
relationship between two entities rather than two concepts. Mathematical
proof based on computer enumeration exists. Yet, it does not have the
ability to infer from one concept to another. 

To narrow the gap, we may construct the concept (or rule) graph whose
nodes represent different concepts. For example, from two concepts ``a
car runs faster than a horse'' and ``a horse runs faster than a man'',
we infer that ``a car runs faster than a man'' through the transitive
law. A concept graph in a general domain could be too complex to build.
Yet, it could be feasible to do it in a special domain. For example, if
we focus on ``I for health care'', the number of concepts is much
smaller.  There is difference between the concept graph and the
traditional expert system. An expert system does not have links between
nodes while links in the concept graph introduce the logical
relationship between two concepts. Common sense reasoning and rules
discovery/creation are possible through inductive learning, i.e., path
finding in domain-specific concept graphs. 

Humans react to external stimuli with responses such as action, decision
and planning.  Rational responses are knowledge-based.  Action/decision
is often related to penalties and/or rewards.  Reinforcement learning is
developed with this principle via cost function definition and
optimization. This proves to be effective in gaming (e.g., chess and
go). Game theory offers an alternative optimal decision process among
independent and competing actors in strategic settings. 

Generalization of reinforcement learning and game theory to real-world
situations is however non-trivial since it is difficult to define proper
cost functions. Furthermore, human behavior involves intuition,
instinct, psychological factors and constraints (e.g. faith and ethics),
which are difficult to model. For the next generation AI to be fully
autonomous, we need a clearly defined goal. For example, medical
diagnosis can be conducted by AI automatically while medical treatment
will be determined by AI and humans jointly since the latter involves
human factors. Also, it is the human who should take the ultimate
responsibility. 

\section{Conclusion}\label{sec:conclusion}

Resolving the semantic gap is a topic that has attracted growing
attention in the artificial intelligence community. Unlike previous
papers, this survey drew experiences from two fundamental computer
vision problems: object detection and metric learning in image
retrieval. The central theme was on the role of supervision, which was
accomplished by ``data annotation schemes'' and ``design of loss
functions''. We organized the survey by various supervision forms.
Furthermore, we offer a broader perspective on intelligence gaps and
discuss a couple of ideas in resolving these gaps to shed light on
future research directions. 

\bibliographystyle{ieee}
\bibliography{egbib}

\end{document}